\documentclass[twocolumn]{svjour3}

\usepackage{times}
\usepackage{epsfig}
\usepackage{graphicx}
\usepackage{graphics}
\usepackage{amsmath}
\usepackage{amssymb}

\usepackage{comment}
\usepackage{hhline}

\usepackage{caption} 
\usepackage[export]{adjustbox}
\usepackage{tabularx}

\usepackage{url}

\usepackage{enumitem}

\usepackage[pagebackref=true,breaklinks=true,letterpaper=true,colorlinks,bookmarks=false]{hyperref}

\begin{document}

\title{Face Behavior \`a la carte: \\Expressions, Affect and Action Units in a Single Network}

\titlerunning{Face Behavior \`a la carte: \\Expressions, Affect and Action Units in a Single Network}

\author{Dimitrios Kollias $^\star$        \and
        Viktoriia Sharmanska $^\dagger$\and
        Stefanos Zafeiriou$^{2}$
}

\date{Received:  / Accepted: }

\authorrunning{D. Kollias}

\institute{    
$^\star$dimitrios.kollias15@imperial.ac.uk\\
$^\dagger$sharmanska.v@imperial.ac.uk\\
$^{2}$s.zafeiriou@imperial.ac.uk\\
\at
$^{\star,\dagger,2}$Department of Computing, Imperial College London, Queen’s Gate, London SW7 2AZ, UK\\
$^{\star,\dagger,2}$ FaceSoft Ltd \\
$^{2}$Center for Machine Vision and Signal Analysis, University of
Oulu, Oulu, Finland\\
}

\maketitle

\begin{abstract}
Automatic facial behavior analysis has a long history of studies in the intersection of computer vision, physiology and psychology. However it is only recently, with the collection of large-scale datasets and powerful machine learning methods such as deep neural networks, that automatic facial behavior analysis started to thrive. Three of its iconic tasks are automatic recognition of basic expressions (e.g. happiness, sadness, surprise), estimation of continuous affect (e.g., valence and arousal), and detection of facial action units (activations of e.g. upper/inner eyebrows, nose wrinkles). 
Up until now these tasks have been studied independently by collecting a dedicated dataset and training a single-task model. We present the first and the largest study of all facial behaviour tasks learned jointly in a single holistic framework, which we call FaceBehaviorNet. For this we utilize all publicly available datasets in the community (over 5M images) that study facial behaviour tasks in-the-wild. We demonstrate that training jointly an end-to-end network for all tasks  has consistently better performance than training each of the single-task networks. Furthermore, we propose two simple strategies for coupling the tasks during training, co-annotation and distribution matching, and show the advantages of this approach. Finally we show that FaceBehaviorNet has learned features that encapsulate all aspects of facial behaviour, and can be successfully applied to perform tasks (compound emotion recognition) beyond the ones that it has been trained in a zero- and few-shot learning setting. The model and source code will be made publicly available.

\end{abstract}
%\vspace*{-3px}

%%%%%%%%% BODY TEXT
\section{Introduction}

Holistic frameworks, where several parts, e.g. learning tasks, are interconnected and explicable by the reference to the whole, are common in computer vision. The diverse examples range from the scene understanding framework that reasons about 3D object detection, pose estimation, semantic segmentation and depth reconstruction \cite{wang2015holistic}, the face analysis framework that addresses face detection, landmark localization, gender recognition, age estimation \cite{ranjan2017all}, to the universal networks for low-, mid-, high-level vision \cite{kokkinos2017ubernet} and for various visual tasks \cite{zamir2018taskonomy}. Most if not all of these prior works rely on building a multi-task framework where learning is done based on the ground truth annotations with full or partial overlap across tasks. During training, all the tasks are optimised simultaneously aiming for representation learning that supports the holistic view. 

In this work we propose the first holistic framework for emotional behaviour analysis in-the-wild, where different emotional states such as binary action units activations, basic categorical emotions and continuous dimensions of valence and arousal constitute the interconnected tasks that are explicable by the human's affective state. What makes it different from the aforementioned holistic approaches is exploring the idea of task-relatedness, given explicitly either as external expert knowledge or from empirical evidence. In this form, it is similarly motivated to the classical multi-task literature exploring feature sharing \cite{argyriou2007multi} and task relatedness \cite{jayaraman2014decorrelating} during training; more examples can be found in the surveys \cite{zhang2017survey,pan2010survey}. However in the multi-task setting, one typically assumes homogeneity of the tasks, i.e. tasks of the same type such as object classifiers or attribute detectors. The main difference and novelty of our work is that the proposed holistic framework (i) explores the relatedness of non-homogeneous tasks, e.g. tasks for classification, detection, regression; (ii) operates over datasets with partial or non-overlapping annotations of the tasks; (iii) encodes explicit relationship between tasks to improve transparency and to enable expert input.

Recently, a lot of effort has been made towards collecting large scale datasets of naturalistic behaviour captured in uncontrolled conditions, \emph{in-the-wild} \cite{kollias2018deep,zafeiriou2017aff,mollahosseini2017affectnet,emotionet2016}, which is the focus of our study.  
There is a rich literature on recognition of basic emotion categories or expressions \cite{ekman1971constants} such as anger, disgust, fear, happiness, sadness, surprise and neutral in-the-wild \cite{dalgleish2000handbook,cowie2003describing}. Continuous affect dimensions such as valence (how positive/negative a person is) and arousal (how active/passive a person is) have attracted attention (VA) recently, as they are naturally suited to represent emotional state and its changes over time. Datasets for continuous affect are simpler to collect while benefiting from human computer interactions techniques. 
Automatic facial analysis has been also studied in terms of the facial action units (AUs) coding system  \cite{ekman1997face}. This system is a systematic way to code the facial motion with respect to activation of facial muscles. It has been widely adopted as a common standard towards systematically categorising physical manifestation of complex facial expressions. The dataset collection of action units is very costly, as it requires skilled annotators to perform the task. Nevertheless there has been a lot of effort to collect action unit annotations and develop automatic AUs annotation toolboxes \cite{emotionet2016,openface2015}. 

Up until now facial behaviour in-the-wild has been primarily addressed by collecting in-the-wild datasets to solve individual tasks. However the three aforementioned tasks of facial behaviour analysis are interconnected. In \cite{ekman1997face}, the facial action coding system (FACS) has been built to indicate for each of the basic expressions its prototypical action units. In \cite{du2014compound}, a dedicated user study has been conducted to study the relationship between AUs activations and emotion expressions beyond basic types -- compound emotions (e.g. happily surprised). 
In \cite{khorrami2015deep}, the authors show that neural networks trained for expression recognition implicitly learn facial action units. 

Also, in \cite{mehu2015emotion} the authors have discovered that valence and arousal dimensions could be interpreted by AUs. For example, AU12 (lip corner puller) is related to positive valence.

Our main contributions are as follows: 
\begin{itemize}[leftmargin=*,noitemsep,nolistsep]
\item We propose a flexible holistic framework that can accommodate non-homogeneous tasks with encoding prior knowledge of tasks relatedness. In our experiments we evaluate two effective strategies of task relatedness: 
%\footnote{Other scenarios, e.g. face analysis with tasks of identity recognition and \textcolor{red}{facial attributes predictions a.k.a basic emotions and action units}, will be assessed in the extended version of this manuscript.}: 
a) obtained from a cognitive and psychological study, e.g. how action units are related to basic emotion categories \cite{du2014compound}, and b) inferred empirically from external dataset annotations; the annotations will be made publicly available.
\item 
We propose an effective algorithmic approach of coupling the tasks via co-annotation and distribution matching and show its effectiveness for facial behaviour analysis; 
\item We present the first, to the best of our knowledge, holistic network for facial behaviour analysis (FaceBehaviorNet) and train it end-to-end for predicting simultaneously 7 basic expressions, 17 action units and  continuous valence-arousal in-the-wild. For network training we utilize all publicly available in-the-wild databases that, in total, consist of over 5M images with partial and/or non-overlapping annotations for different tasks. 
%We evaluate two sources of task relatedness: 
%\footnote{Other scenarios, e.g. face analysis with tasks of identity recognition and \textcolor{red}{facial attributes predictions a.k.a basic emotions and action units}, will be assessed in the extended version of this manuscript.}: 
%a) obtained from a cognitive and psychological study, e.g. how action units are related to basic emotion categories \cite{du2014compound}, and b) inferred empirically from external dataset annotations. 
\item We show that FaceBehaviorNet greatly outperforms each of the single-task networks, validating that our network's emotion recognition capabilities are enhanced when it is jointly trained for all related tasks.
We further explored the feature representation learned in the joint training and show its generalization abilities on the task of compound expressions recognition when no or little training data is available (zero-shot and few-shot learning). 
\end{itemize}

\section{Related work}\label{related_work}

Works exist in literature that use emotion labels to complement missing AU annotations or increase generalization of AU classifiers \cite{ruiz2015emotions,yang2016multiple,wang2017expression}. Our work deviates from such methods, as we target a joint learning of three facial behaviour tasks via a single holistic framework, whilst these works perform  only AU detection and not emotion recognition (nor valence-arousal estimation).

Multi-task learning (MTL) was first studied in \cite{caruana1997multitask}, where the authors propose to jointly learn parallel tasks sharing a common representation and transferring part of the knowledge learned to solve one task to improve the learning of the other related tasks. Since then, several approaches have adopted MTL for solving different problems in computer vision and machine learning. In the face analysis domain,
the use of MTL is somewhat limited. In \cite{wang2017multi}, MTL was tackled through a neural network that jointly handled face recognition and facial attribute prediction tasks. MTL helped capture global feature and local attribute information simultaneously.

One of the closest goals to ours is \cite{weichi}, where an integrated deep learning framework (FATAUVA-Net) for sequential facial attribute recognition, AU detection, and valence-arousal estimation was proposed. This framework employed face attributes as low-level (first component) and AUs as mid-level (second component) representations for predicting quantized valence-arousal values (third component). However training of this model is made of transfer learning and fine-tuning steps, is hierarchical and not end-to-end. 
In a similar work of \cite{wang2018two}, a two-level attention with two stage multi-task learning framework was constructed for emotion recognition and valence-arousal estimation; this work was based on a database (AffectNet \cite{mollahosseini2017affectnet}) annotated for both tasks. In the first attention level, a CNN extracted position-level features and then in the second an RNN with self-attention was proposed to model the relationship between layer-level features.

\section{The Proposed Approach}\label{approach}
We start with the multi-task formulation of the facial behaviour model. 
In this model we have three objectives: (1) learning seven basic emotions, 
(2) detecting activations of $17$ binary facial action units, (3) learning the intensity of the valence and arousal continuous affect dimensions. 
We train a multi-task neural network model to jointly perform (1)-(3). 
For a given image $x \in \mathcal{X}$, we can have label annotations of either one of seven basic emotions $y_{emo} \in \{1,2,\ldots,7\}$, or $17$\footnote{In fact, $17$ is an aggregate of action units in all datasets; typically each dataset has from 10 to 12 AUs labelled by purposely trained annotators.} binary action units activations $y_{au} \in \{0,1\}^{17}$, or two continuous affect dimensions, valence and arousal, $y_{va} \in [-1,1]^{2}$.
For simplicity of presentation, we use the same notation $x$ for all images leaving the context to be explained by the label notations.  
We train the multi-task model by minimizing the following objective:
%\vspace*{-4px}
\begin{align}
\mathcal{L}_{MT} &= \mathcal{L}_{Emo} + \lambda_{1} \mathcal{L}_{AU} + \lambda_{2} \mathcal{L}_{VA} \label{eq:mt1}\\
\mathcal{L}_{Emo} &= \mathbb{E}_{x,y_{emo}}[-\text{log } p (y_{emo}|x)]\nonumber\\ 
\mathcal{L}_{AU} &= \mathbb{E}_{x,y_{au}}[- \text{log } p (y_{au}|x)]\nonumber\\ 
\mathcal{L}_{VA} &= 1- CCC(y_{va},\bar{y}_{va}),\nonumber
\end{align}
where the first term is the cross entropy loss computed over images with a basic emotion label, the second term is the binary cross entropy loss computed over images with $17$ AUs activations, $\text{log } p (y_{au}|x) := [ \sum_{k=1}^{17} \delta_k]^{-1}  \cdot \sum_{i=1}^{17} \delta_i \cdot$

\smallskip \noindent
$[y_{au}^i\text{log } p (y_{au}^i|x) + (1-y_{au}^i)\text{log } (1-p (y_{au}^i|x))]$, 
where $\delta_i \in \{0,1\}$ indicates whether the image contains annotation for $AU_i$.
The third term measures the concordance correlation coefficient between the ground truth valence and arousal $y_{va}$ and the predicted $\bar{y}_{va}$, $CCC(y_{va},\bar{y}_{va})=\frac{\rho_a + \rho_v}{2}$, where for $i \in \{v,a\}$, $y_i$ is the ground truth, $\bar{y_i}$ is the predicted value and  $\rho_i=$
\begin{align}
 \frac{2\cdot\mathbb{E}[(y_{i}-\mathbb{E}_{y_{i}})\cdot(\bar{y}_{i}-\mathbb{E}_{\bar{y}_{i}})]}
{\mathbb{E}^2[{(y_{i}-\mathbb{E}_{y_{i}}})^2] + \mathbb{E}^2[{(\bar{y}_{i}-\mathbb{E}_{\bar{y}_{i}}})^2] + (\mathbb{E}_{y_{i}} -\mathbb{E}_{\bar{y}_{i}})^2}.\nonumber
\label{eq:ccc}
\end{align}

%\vspace{-0.4cm}
\paragraph{Coupling of basic emotions and AUs via co-annotation}
In the seminal work \cite{du2014compound}, the authors conduct a study on the relationship between emotions (basic and compound) and facial action units activations. The summary of the study is a table of the emotions and their prototypical and observational actions units (Table 1 in \cite{du2014compound}) which we include in Table \ref{table:EmoAUs} for completeness. Prototypical are action units that are labelled as activated across all annotators' responses, observational are action units that are labelled as activated by a fraction of annotators. For example, in emotion \emph{happiness} the prototypical are AU12 and AU25, the observational is AU6 with weight $0.51$ (observed by 51\% of the annotators). 

Here let us mention that Table \ref{table:EmoAUs} constitutes the relatedness between the emotion categories and action units obtained from a cognitive study. In our experiments, in Section \ref{empirical}, we also show that such relatedness can be inferred empirically from external dataset annotations. Other means of describing task relatedness in a holistic framework will be further explored in the future.

We propose a simple strategy of \emph{co-annotation} to couple the training of emotions and action unit predictions. 
Given an image $x$ with the ground truth %annotation of 
basic emotion $y_{emo}$, we enforce the prototypical and observational AUs of this emotion to be activated. We co-annotate the image $(x,y_{emo})$ with $y_{au}$; %that contains only the prototypical and observational AUs
 this image contributes to both 
%and include this image twice, when computing
$\mathcal{L}_{Emo}$ and $\mathcal{L}_{AU}$\footnote{Here we overload slightly our notations; for co-annotated images, $y_{au}$ has variable length and only contains prototypical and observational AUs.} in eq.  \ref{eq:mt1}. We re-weight the contributions of the observational AUs with the annotators' agreement score (from Table \ref{table:EmoAUs}).

Similarly, for an image $x$ with the ground truth  %annotation of the 
action units $y_{au}$, we check whether we can co-annotate it with an emotion label. 
{For an emotion to be present, all its prototypical and observational AUs have to be present. In cases when more than one emotion is possible, we assign the label $y_{emo}$ of the emotion with the largest requirement of prototypical and observational AUs}.
The image $(x,y_{au})$ that is co-annotated with the emotion label $y_{emo}$ %is included twice in \ref{eq:mt1}, when computing
contributes to both $\mathcal{L}_{AU}$ and $\mathcal{L}_{Emo}$ in eq. \ref{eq:mt1}. 
We call this approach the FaceBehaviorNet with co-annotation. 

\begin{table}[t]
\caption{Basic emotions and their prototypical and observational AUs from \cite{du2014compound}. The weights $w$ in brackets correspond to the fraction of annotators that observed the AU activation.}
\label{table:EmoAUs}
\centering
\scalebox{1.}{
\begin{tabular}{|l|c|c|}
\hline
Emotion   & Protot. AUs & Observ. AUs (with weights $w$)\\
\hline\hline
happiness &  12, 25 & 6 (0.51) \\
\hline
sadness &  4, 15 & 1 (0.6), 6 (0.5), 11 (0.26), 17 (0.67) \\
\hline
fear &  1, 4, 20, 25 &2 (0.57), 5 (0.63), 26 (0.33) \\
\hline
anger &4, 7, 24 &10 (0.26), 17 (0.52), 23 (0.29)\\
\hline
surprise &1, 2, 25, 26 &5 (0.66)\\
\hline
disguste &9, 10, 17 & 4 (0.31), 24 (0.26)\\
\hline
\end{tabular}
}
\end{table}

\paragraph{Coupling of basic emotions and AUs via distribution matching} 
The aim here is to align the \emph{predictions} of the emotions and action units tasks during training.  
For each sample $x$ we have the predictions of emotions $p(y_{emo}|x)$ as the softmax scores over seven basic emotions and we have the prediction of AUs activations $p(y_{au}^i|x)$, $i=1,\ldots,17$ as the sigmoid scores over $17$ AUs. 

The distribution matching idea is simple: we match the distribution over AU predictions $p(y_{au}^i|x)$ with the distribution 
$q(y_{au}^i|x)$, where the AUs are modeled as a mixture over the basic emotion categories: %the prototypical and observational AUs: 
\begin{equation}
    q(y_{au}^i|x) = \sum_{y_{emo} \in \{1,\ldots,7\}} p(y_{emo}|x) \: p(y_{au}^i| y_{emo}), %{\sum_{y_{emo} \in \{1,\ldots,7\}} \: p(y_{au}^i| y_{emo})}.
\label{eq:distr}
\end{equation} 
where $p(y_{au}^i| y_{emo})$ is defined deterministically from Table~\ref{table:EmoAUs} %: $p(y_{au}^i| y_{emo})=1$ 
and is 1 for prototypical/observational action units, or 0 otherwise.  For example, AU2 is prototypical for emotion \emph{surprise} and observational for emotion \emph{fear} and thus $q(y_{\text{AU2}}|x) = \frac{1}{2}(p(y_{\text{surprise}}|x) + p(y_{\text{fear}}|x))$\footnote{We also tried a variant with reweighting for observational AUs, i.e. $p(y_{au}^i| y_{emo})=w$ }.  

This matching aims to make the network's predicted AUs consistent with the prototypical and observational AUs of the network's predicted emotions. So if, e.g., the network predicts the emotion \emph{happiness} with probability 1,
i.e., \\ $p(y_{\text{happiness}}|x)=1$, then the prototypical and observational AUs of \emph{happiness} -AUs 12, 25 and 6- need to be activated in the distribution q: $q(y_{\text{AU12}}|x) = q(y_{\text{AU25}}|x) = q(y_{\text{AU6}}|x) = 1$; $q(y_{au}^i|x) = 0$, $i \in \{1,..,14\}$. %Since the network predicts that someone is \emph{happy}, we have a belief that AUs 6, 12 and 25 could, with the above probabilistic confidence, be activated. The loss aligns this belief with the corresponding network’s AU predictions.

In spirit of the distillation approach \cite{hinton2015distilling}, we match the distributions $p(y_{au}^i|x)$ and $q(y_{au}^i|x)$ 
by minimizing the cross entropy with the soft targets loss term\footnote{This can be seen as minimizing the KL-divergence $KL(p||q)$ across the $17$ action units.}:
\begin{align}
\mathcal{L}_{DM} = \mathbb{E}_{x} \sum_{i=1}^{17}[ -p(y_{au}^i|x)\text{log }q(y_{au}^i|x) ], \label{eq:coupleloss}
\end{align}
where all available training samples are used to match the predictions.
We call this approach FaceBehaviorNet with distr-matching.

\smallskip
A mix of the two strategies, co-annotation and distribution matching, is also possible. Given an image $x$ with the ground truth annotation of the action units $y_{au}$, we can first co-annotate it with a \emph{soft label} in form of the  distribution over emotions and then match it with the predictions of emotions $p(y_{emo}|x)$. 
More specifically, for each basic emotion, we compute the score over its prototypical and observational AUs being present. For example, for emotion \emph{happiness}, we compute $(y_{\text{AU12}} + y_{\text{AU25}} + 0.51 \cdot y_{\text{AU6}}) / (1+1+0.51)$, or all weights equal $1$ if without reweighting. We take a softmax over the scores to produce the probabilities over emotion categories. In this variant, every single image that has ground truth annotation of AUs will have a \emph{soft} emotion label assigned. Finally we match the predictions $p(y_{emo}|x)$ and the soft label by minimizing the cross entropy with the soft targets similarly to eq. \ref{eq:coupleloss}. We call this approach FaceBehaviorNet with soft co-annotation.

\paragraph{Coupling of categorical emotions, AUs with continuous affect} 
In our work, continuous affect (valence and arousal) is implicitly coupled with the basic expressions and action units via a joint training procedure. Also one of the datasets we used has annotations for categorical and continuous emotions (AffectNet \cite{mollahosseini2017affectnet}). Studying an explicit relationship between them is a novel research direction beyond the scope of this work.

\begin{figure*}[t]
\centering
\adjincludegraphics[height=9cm,width=11cm]{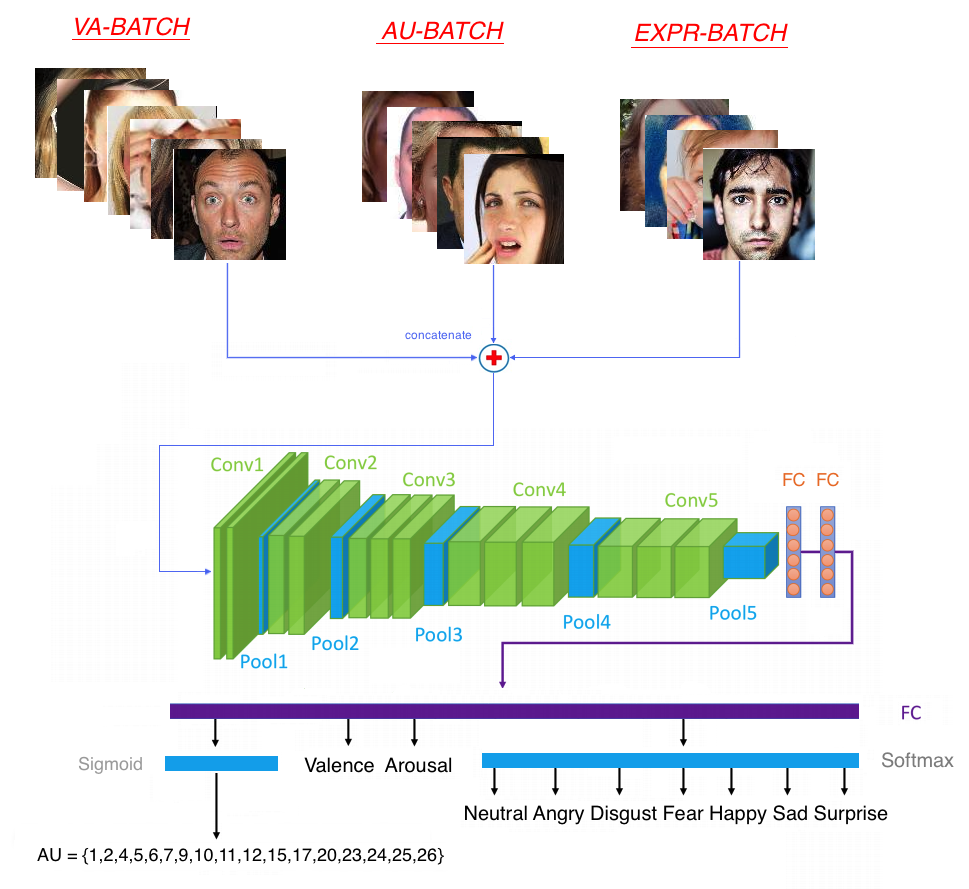} 
\caption{The holistic (multi-task, multi-domain, multi-label) FaceBehaviorNet; 'VA/AU/EXPR-BATCH' refers to batches annotated in terms of VA/AU/7 basic expressions} 
\label{facebehaviornet}
\end{figure*}

\paragraph{FaceBehaviorNet structure}
Fig.\ref{facebehaviornet} shows the structure of the  holistic (multi-task, multi-domain and multi-label) FaceBehaviorNet, based on the 13 convolutional and pooling layers of VGG-FACE \cite{parkhi2015deep} (its fully connected layers are discarded), followed by 2 fully connected layers, each with 4096 hidden units. A (linear) output layer follows that gives final estimates for valence and arousal; it also gives 7 basic expression logits that are passed through a softmax function to get the final 7 basic expression predictions; lastly, it gives 17 AU logits that are passed through a sigmoid function to get the final 17 AU predictions. One can see that the predictions  for  all  tasks  are  pooled  from  the  same  feature  space. 

\section{Experimental Study}

\paragraph{Databases}

Let us first describe the databases that we utilized in all our experiments. We selected to work with these databases because they provide a large number of samples with accurate annotations of valence-arousal, basic expressions and AUs. Training with these datasets allows our networks to learn to recognize affective states under a large number of image conditions (e.g., each database includes images at different resolutions, poses, orientations and lighting conditions). These datasets also include a variety of samples in both genders, ethnicities and races.

The \textbf{Aff-Wild} database \cite{kollias2018deep} \cite{zafeiriou2017aff} has been the first large scale captured in-the-wild database, containing 298 videos (200 subjects) of around 1.25M frames, annotated in terms of valence-arousal. It served as benchmark for the Aff-Wild Challenge organized in CVPR 2017.
%%%%%%%%%%%%%%%%%%%%%%%%%%%%%%%%%%%%%%%%%%%%%%%%%%%%%%
%%%%%%%%%%%%%%%%%%%%%%%%%%%%%%%%%%%%%%%%%%%%%%%%%%%%%%
The \textbf{AffectNet} database \cite{mollahosseini2017affectnet} contains around 1M facial images, 400K of which were manually annotated in terms of 7 discrete expressions (plus contempt) and valence-arousal. 
%%%%%%%%%%%%%%%%%%%%%%%%%%%%%%%%%%%%%%%%%%%%%%%%%%%%%%
%%%%%%%%%%%%%%%%%%%%%%%%%%%%%%%%%%%%%%%%%%%%%%%%%%%%%%
The \textbf{AFEW} database\cite{dhall2017individual} is used in the EmotiW Challenges that focus on audiovisual classification of each of the 1,809 video clips into the 7 basic emotion categories. 
%%%%%%%%%%%%%%%%%%%%%%%%%%%%%%%%%%%%%%%%%%%%%%%%%%%%%%
%%%%%%%%%%%%%%%%%%%%%%%%%%%%%%%%%%%%%%%%%%%%%%%%%%%%%%
The \textbf{RAF-DB} database \cite{li2017reliable} contains 15.2K facial images annotated in terms of the 7 basic and 11 compound emotion categories. 

%%%%%%%%%%%%%%%%%%%%%%%%%%%%%%%%%%%%%%%%%%%%%%%%%%%%%%
%%%%%%%%%%%%%%%%%%%%%%%%%%%%%%%%%%%%%%%%%%%%%%%%%%%%%%
The \textbf{EmotioNet} database \cite{fabian2016emotionet} is a large-scale database with around 1M facial expression images; 950K images were automatically annotated and the remaining 50K images were manually annotated with 11 AUs. Additionally, a subset of about 2.5K images was annotated with the 6 basic and 10 compound emotions. It was released for the EmotioNet Challenge in 2017 \cite{benitez2017emotionet}. 
%%%%%%%%%%%%%%%%%%%%%%%%%%%%%%%%%%%%%%%%%%%%%%%%%%%%%%
%%%%%%%%%%%%%%%%%%%%%%%%%%%%%%%%%%%%%%%%%%%%%%%%%%%%%%
The \textbf{DISFA} database \cite{mavadati2013disfa} is a lab controlled database with spontaneous emotion expressions, annotated for the presence, absence and intensity of 12 AUs. It consists of 260K video frames of 27 subjects recorded by two cameras. 
%%%%%%%%%%%%%%%%%%%%%%%%%%%%%%%%%%%%%%%%%%%%%%%%%%%%%%
%%%%%%%%%%%%%%%%%%%%%%%%%%%%%%%%%%%%%%%%%%%%%%%%%%%%%%
The \textbf{BP4D-Spontaneous} database\cite{zhang14bp4d} (in the rest of the paper we refer to it as BP4D) contains 61 subjects with 223K frames and is annotated for the occurrence and intensity of 27 AUs. It has been used as a part of the FERA 2015 Challenge \cite{valstar2015fera}.
%%%%%%%%%%%%%%%%%%%%%%%%%%%%%%%%%%%%%%%%%%%%%%%%%%%%%%
%%%%%%%%%%%%%%%%%%%%%%%%%%%%%%%%%%%%%%%%%%%%%%%%%%%%%%
The \textbf{BP4D+} database \cite{zhang2016multimodal} is an extension of BP4D incorporating different modalities as well as more subjects (140). It is annotated for occurrence of 34 AUs and intensity for 5 of them. It has been used as a part of the FERA 2017 Challenge \cite{valstar2017fera}.

Here let us note that for AffectNet, AFEW, BP4D and BP4D+, no test set is released; thus we use the released validation set to test on and randomly divide the training set into a training and a validation subset (with a 85/15 split).

%\vspace*{-0.5cm}
\paragraph{Performance Measures}
We use: i) the CCC for Aff-Wild (CCC was the evaluation criterion of Aff-Wild Challenge) and Affectnet, ii) the total accuracy for AFEW (this metric was the evaluation criterion of the EmotiW Challenges), the mean diagonal value of the confusion matrix for RAF-DB (this criterion was selected for evaluating the performance on this database by \cite{li2017reliable}), the F1 score for AffectNet, iii) the F1 score for DISFA, BP4D and BP4D+ (this metric was the evaluation criterion of the FERA 2015 and 2017 Challenges); for AU detection in EmotioNet the Challenge's metric was the average between: a) the mean (across all AUs) F1 score and b) the mean (across all AUs) accuracy; for the expression classification, it was the average between:  a) the mean (across all emotions) F1 score and b) the unweighted average recall (UAR) over all emotion categories.

%\vspace*{-0.5cm}
\paragraph{Pre-Processing}

We used the SSH detector \cite{najibi2017ssh} based on ResNet and trained on the WiderFace dataset \cite{yang2016wider} to extract, from all images, face bounding boxes and 5 facial landmarks; the latter were used  for face alignment. All   cropped   and   aligned   images   were  resized   to $96 \times 96 \times 3$ pixel resolution and their intensity values were normalized  to  $[-1,1]$.  %Those  images  were used as inputs for training our networks.

\subsection{Training Implementation Details}\label{training}

At this point let us describe the strategy that was used for feeding images from different databases to FaceBehaviorNet. At first, the training set was split into three different sets, each of which contained images that were annotated in terms of either valence-arousal, or action units, or seven basic expressions; let us denote these sets as VA-Set, AU-Set and EXPR-Set, respectively. During training, at each iteration, three batches, one from each of these sets (as can be seen in Fig.\ref{facebehaviornet}), were concatenated and fed to FaceBehaviorNet. This step is important for network training, because: i) the network minimizes the objective function of eq. \ref{eq:mt1}; at each iteration, the network has seen images from all categories and thus all loss terms contribute to the objective function, ii) since the network sees an adequate number of images from all categories, the weight updates (during gradient descent) are not based on noisy gradients; this in turn prevents poor convergence behaviors; otherwise, we would need to tackle these problems, e.g. do asynchronous SGD as proposed in \cite{kokkinos2017ubernet} to make the task parameter updates decoupled, 
iii) the CCC cost function (defined in Section \ref{approach}) needs an adequate sequence of predictions.

Since VA-Set, AU-Set and EXPR-Set had different sizes, they needed to be  'aligned'. To do so, we selected the batches of these sets in such a manner, so that after one epoch we will have sampled all images in the sets. In particular, we chose batches of size 401, 247 and 103 for the VA-Set, AU-Set and EXPR-Set, respectively. 
The training of FaceBehaviorNet was performed in an end-to-end manner, with a learning rate of $10^{-4}$. A 0.5 Dropout value was used in the fully connected layers. Training was performed on a Tesla V100 32GB GPU; training time was about 2 days. %The Tensorflow platform has been used. 

\subsection{Task-Relatedness from Empirical Evidences}\label{empirical}

Table \ref{table:EmoAUs} was created using a cognitive and psychological study with human participants. Here, we create another Table inferred empirically from external dataset annotations. In particular, we use the recently proposed Aff-Wild2 database \cite{kollias2019expression,kollias2018aff2,kollias2018multi22,kollias2020analysing}, which is the first  in-the-wild database that contains annotations for all three behavior tasks that we are dealing with in this paper. It consists of 558 videos: all contain VA annotations, 63 contain AU annotations and 84 contain basic expression annotations. It served as benchmark for the ABAW Competition organized in IEEE FG 2020.

\begin{table*}[ht]
\caption{Performance evaluation of valence-arousal, seven basic expression and action units predictions on all used databases provided by the FaceBehaviorNet when trained with/without the coupled losses, under the two task relatedness scenarios. }
\label{comparison_losses}
\centering
\scalebox{0.8}{
\begin{tabular}{ |c||c||c|c|c|c|c|c|c|c|c|c|c|c| }
 \hline
\multicolumn{1}{|c||}{\begin{tabular}{@{}c@{}} Databases \end{tabular}} & \multicolumn{1}{c||}{\begin{tabular}{@{}c@{}} Relatedness \end{tabular}}  & \multicolumn{2}{c|}{Aff-Wild} & \multicolumn{3}{c|}{\begin{tabular}{@{}c@{}}  AffectNet \end{tabular}} & \multicolumn{1}{c|}{AFEW}  & \multicolumn{1}{c|}{RAF-DB} & \multicolumn{2}{c|}{EmotioNet} & \multicolumn{1}{c|}{DISFA}  & \multicolumn{1}{c|}{BP4D} & \multicolumn{1}{c|}{BP4D+}  \\
 \hline
 FaceBehaviorNet &   &\multicolumn{1}{c|}{CCC-V}  &\multicolumn{1}{c|}{CCC-A} &\multicolumn{1}{c|}{CCC-V} &\multicolumn{1}{c|}{CCC-A} &\multicolumn{1}{c|}{\begin{tabular}{@{}c@{}}  F1 \\ Score \end{tabular}}  & \multicolumn{1}{c|}{\begin{tabular}{@{}c@{}}  Total \\ Accuracy \end{tabular}} & \multicolumn{1}{c|}{\begin{tabular}{@{}c@{}}  Mean diag. \\ of conf. matrix \end{tabular}} &\multicolumn{1}{c|}{\begin{tabular}{@{}c@{}}  F1 \\ Score \end{tabular}} &  \multicolumn{1}{c|}{\begin{tabular}{@{}c@{}} Accuracy \end{tabular}} & \multicolumn{1}{c|}{\begin{tabular}{@{}c@{}}  F1 \\ Score \end{tabular}}&\multicolumn{1}{c|}{\begin{tabular}{@{}c@{}}  F1 \\ Score \end{tabular}} &\multicolumn{1}{c|}{\begin{tabular}{@{}c@{}}  F1 \\ Score \end{tabular}} \\ 
  \hhline{=:=:=:=:=:=:=:==:=:=:=:=:=}
no coupling loss & -  & 0.55 & 0.36 & 0.56 & 0.46 & 0.54 & 0.38 & 0.67 & 0.49 & 0.94 & 0.52 & 0.61 & 0.57  \\
 \hline
 \hline
co-annotation & \cite{du2014compound} & 0.56  & 0.38  & 0.56  & 0.46 & 0.55 & 0.40  & 0.67  & 0.49  & 0.94   &  0.54 & 0.64 & 0.58  \\
 \hline
soft co-annotation & \cite{du2014compound}  & 0.56 & 0.39 & 0.57  & 0.47 & 0.57  & 0.41  & 0.67  & 0.50  & 0.94  & 0.54  & 0.64 & 0.60  \\
 \hline
 distr-matching & \cite{du2014compound}  & 0.56  & 0.37 & 0.57  & 0.49 & 0.57  & 0.42  & 0.68  & 0.50  & 0.94   & 0.56 & 0.66  & 0.58  \\ 
 \hline 
\begin{tabular}{@{}c@{}} \textbf{soft co-annotation} \\ \textbf{and distr-matching}  \end{tabular} & \cite{du2014compound} & 0.59  & \textbf{0.41}  & 0.59 & 0.50  & \textbf{0.60}   & \textbf{0.43}    & 0.70  & 0.51  & \textbf{0.95}   & 0.57  & \textbf{0.67}  & \textbf{0.60}  \\
 \hline 
 \hline
co-annotation & Aff-Wild2 & 0.55  & 0.37  & 0.56  & 0.47 & 0.54 & 0.40  & 0.67  & 0.50  & 0.93 &  0.54 & 0.61 & 0.57  \\
\hline
soft co-annotation & Aff-Wild2  & 0.56 & 0.37 & 0.57  & 0.47 & 0.55  & 0.42  & 0.68  & 0.52  & 0.94  & 0.58  & 0.63 & 0.59  \\
 \hline
 distr-matching & Aff-Wild2  & 0.57  & 0.39 & 0.60  & \textbf{0.51} & 0.57  & 0.42  & 0.69  & 0.50  & 0.94 & 0.57 & 0.62  & 0.58  \\ 
 \hline 
\begin{tabular}{@{}c@{}} \textbf{soft co-annotation} \\ \textbf{and distr-matching}  \end{tabular} &  Aff-Wild2 & \textbf{0.60}  & 0.40  & \textbf{0.61} & \textbf{0.51}  & \textbf{0.60}   & 0.42    & \textbf{0.71}  & \textbf{0.54}  & 0.94   & \textbf{0.60}  & 0.66  & \textbf{0.60}  \\ 
 \hline
\end{tabular}
}
\end{table*}

At first, we trained a network for AU detection on the union of Aff-Wild2 and GFT databases \cite{girard2017sayette}. Next, this network was used for automatically annotating all Aff-Wild2 videos with AUs. These annotations will be made  publicly available. Table \ref{table:EmoAUs_Affwild2} shows the distribution of AUs for each basic expression. In parenthesis next to each AU (e.g. AU12) is the percentage of images (0.82) annotated with the specific expression (happiness) in which this AU (AU12) was activated.

%%% relatedness or relationship
\begin{table}[h]
\caption{Relatedness between basic emotions and AUs, inferred from Aff-Wild2.}
\label{table:EmoAUs_Affwild2}
\centering
\scalebox{1.}{
\begin{tabular}{|l|c|}
\hline
Emotion   &  AUs (with weights $w$)\\
\hline\hline
happy &  12 (0.82), 25 (0.7), 6 (0.57), 7 (0.83), 10 (0.63) \\
\hline
sad &  4 (0.53), 15 (0.42), 1 (0.31), 7 (0.13), 17 (0.1) \\
\hline
fearful &  1 (0.52), 4 (0.4), 25 (0.85), 5 (0.38), 7 (0.57), 10 (0.57) \\
\hline
angry & 4 (0.65), 7 (0.45), 25 (0.4), 10 (0.33), 9 (0.15)\\
\hline
surprised &1 (0.38), 2 (0.37), 25 (0.85), 26 (0.3), 5 (0.5), 7 (0.2)\\
\hline
disgusted & 9 (0.21), 10 (0.85), 17 (0.23), 4 (0.6), 7 (0.75), 25 (0.8)\\
\hline
\end{tabular}
}
\end{table}

\subsection{Results: Ablation Study}

At first, we compare the performance of FaceBehaviorNet when trained: i) with only the losses of eq. \ref{eq:mt1} and without using the coupling losses described in Section \ref{approach}, ii) with co-annotation coupling loss, iii) with soft co-annotation coupling loss and iv) with distr-matching coupling loss, vi) with soft co-annotation and distr-matching coupling losses. Table \ref{comparison_losses} shows the results for all these approaches, when Tables \ref{table:EmoAUs} and \ref{table:EmoAUs_Affwild2} are used for the task relatedness. 

Many deductions can be made. 
\textit{Firstly}, when FaceBehaviorNet is trained with any coupling loss, or any combination of these, it displays a better (or in the worst case equal) performance on all databases, in both different task relatedness scenarios. This validates the fact that the proposed losses help to couple the three studied tasks regardless of which relatedness scenario was followed; this shows the generality of the proposed losses that boosted the performance of the network.
\textit{Secondly}, the performance in estimation of valence and arousal improved, although we did not explicitly designed a coupling loss for this; we only coupled emotion categories and action units. We conjecture that when action unit detection and expression classification accuracy is improving (due to coupling), valence and arousal performance also improves, because valence and arousal are implicitly coupled with emotions via joint dataset annotations for both emotion types.

\textit{Thirdly}, in all scenarios, the co-annotation loss results in FaceBehaviorNet having the worst performance when compared to all other coupling losses. 
\textit{Furthermore}, in both settings, when the network was trained with the soft co-annotation loss, the performance increase in AUs was bigger than the corresponding increase in expressions, whereas when the network was trained with the distr-matching loss the performance increase in expressions was bigger than the corresponding increase in AUs.
\textit{Finally}, overall best results have been achieved, in both scenarios, when FaceBehaviorNet was trained with both soft co-annotation and distr-matching losses. In particular, in both settings, an average performance increase of more than 2\% has been observed when using both coupling losses, compared to the (two) cases when only one of them was used.

%% no comparison between settings, just comment on ablation study on losses + general deductions about the losses that are in both settings

\subsection{Results: Comparison with State-of-the-Art and Single-Task Methods}

%%%%%%%%%%%%%%%%%%%%%%%%%%
%%%%%%%%%%%%%%%%%%%%%%%%%%
%%%%%%%%%%%%%%%%%%%%%%%%%
% IF ANY REVIEWER SAYS ABOUT THE NOVELTY THE ANSWER IS:
% WE DO NOT CLAIM THAT WE DEVISED A NEW ARCHITECTURE THAT DOES ALL THESE BLA BLA BLA
% BUT OUR APPROACH IS A GENERAL ONE AND CAN BE APPLIED WITH DIFFERENT ARCHITECTURES + PERFORM + SAY EXPERIMENTS-RESULTS WITH THE COUPLING LOSSES FOR RESNET-101, DESNET-321 (THE BIGGEST ONES)
%%%%%%%%%%%%%%%%%%%%%%%%%%
%%%%%%%%%%%%%%%%%%%%%%%%%%
%%%%%%%%%%%%%%%%%%%%%%%%%

\begin{table*}[ht]
\caption{Performance evaluation of valence-arousal, seven basic expression and action units predictions on all utilized databases provided by the FaceBehaviorNet and state-of-the-art methods.} %on Aff-Wild and AffectNet databases}
\label{comparison_sota}
\centering
\scalebox{0.8}{
\begin{tabular}{ |c||c|c|c|c|c|c|c|c|c|c|c|c| }
 \hline
\multicolumn{1}{|c||}{\begin{tabular}{@{}c@{}} Databases  \end{tabular}} & 
\multicolumn{2}{c|}{Aff-Wild} & \multicolumn{3}{c|}{\begin{tabular}{@{}c@{}}  AffectNet \end{tabular}} & \multicolumn{1}{c|}{AFEW}  & \multicolumn{1}{c|}{RAF-DB} & \multicolumn{2}{c|}{EmotioNet} & \multicolumn{1}{c|}{DISFA}  & \multicolumn{1}{c|}{BP4D} & \multicolumn{1}{c|}{BP4D+}   \\
 \hline
  & CCC-V & CCC-A & CCC-V & CCC-A &\begin{tabular}{@{}c@{}}  F1 \\ Score \end{tabular} & \begin{tabular}{@{}c@{}}  Total \\ Accuracy \end{tabular} & \begin{tabular}{@{}c@{}}  Mean diagonal \\ of conf. matrix \end{tabular}  &\multicolumn{1}{c|}{\begin{tabular}{@{}c@{}}  F1 \\ Score \end{tabular}} &  \multicolumn{1}{c|}{\begin{tabular}{@{}c@{}}  Mean \\ Accuracy \end{tabular}} & \multicolumn{1}{c|}{\begin{tabular}{@{}c@{}}  F1 \\ Score \end{tabular}}&\multicolumn{1}{c|}{\begin{tabular}{@{}c@{}}  F1 \\ Score \end{tabular}} &\multicolumn{1}{c|}{\begin{tabular}{@{}c@{}}  F1 \\ Score \end{tabular}}   \\ 
  \hline
 \hline
best performing CNN\cite{kollias} \cite{kollias2018deep} & 0.51 & 0.33 & - & - & - & - & - & - & - & - &-  & -  \\
\hline 
FATAUVA-Net \cite{weichi}& 0.40 & 0.28 & - & - & - & - & - & - & - & - &-  & -   \\
\hline
(2 $\times$ ) AlexNet \cite{mollahosseini2017affectnet} & - & - & 0.60 & 0.34 & 0.58 & - & - & - & - & - &-  & -   \\ 
 \hline
non-linear SVM\cite{dhall2017individual}& - & - & - &-  & - & 0.38 & - & - & - & - &-  & -   \\
\hline 
VGG-FACE-mSVM\cite{li2017reliable}& - & - & - &-  & - & - & 0.58 & - & - & - &-  & -  \\
\hline 
AlexNet \cite{benitez2017emotionet} & - & - & - &-  & - & - & - & 0.39 & 0.83 & - & - & - \\ 
 \hline
ResNet-34 \cite{ding2017facial} & - & - & - &-  & - & - & -  & \textbf{0.64} & 0.82 & - & - & - \\ 
\hline
DLE extension \cite{yuce2015discriminant} & - & - & - &-  & - & - & -  & - & - & - &0.59 & -  \\
\hline 
\cite{tang2017view} & - & - & - &-  & - & - & -  & - & - & - & -  & 0.58  \\
\hline
\hline
(3 $\times$) VGG-FACE single-task & 0.52 & 0.31 & 0.53 & 0.43 & 0.51 & 0.37 & 0.59 & 0.41 & 0.92 & 0.47 & 0.56 & 0.54 \\
\hhline{=:=:=:=:=:=:=:=:=:=:=:=:=}
\begin{tabular}{@{}c@{}}  FaceBehaviorNet, no coupling loss \end{tabular} & \textit{0.55} & \textit{0.36} & 0.56 & \textit{0.46} & 0.54 & \textit{0.38} & \textit{0.67} & 0.49 & \textit{0.94} & \textit{0.52} & \textit{0.61} & 0.57  \\
\hline
\begin{tabular}{@{}c@{}} \textbf{FaceBehaviorNet, soft co-annotation} \\ \textbf{and distr-matching, \cite{du2014compound}}  \end{tabular} & 0.59  & \textbf{0.41}  & 0.59 & 0.50  & \textbf{0.60}   & \textbf{0.43}    & 0.70  & 0.51  & \textbf{0.95}   & 0.57  & \textbf{0.67}  & \textbf{0.60}  \\
\hline
\begin{tabular}{@{}c@{}} \textbf{FaceBehaviorNet, soft co-annotation} \\ \textbf{and distr-matching, Aff-Wild2}  \end{tabular} & \textbf{0.60}  & 0.40  & \textbf{0.61} & \textbf{0.51}  & \textbf{0.60}   & 0.42    & \textbf{0.71}  & 0.54  & 0.94   & \textbf{0.60}  & 0.66  & \textbf{0.60}  \\ 
\hline
\end{tabular}
}
\end{table*}

%At first, let us mention that we also trained ResNet-50 and DenseNet-121 with all the databases described before and compared their performance to that of FaceBehaviorNet. The latter  provided  the  best  results,  outperforming, on average, the ResNet-50 by 5.33\% and the DenseNet-121 by 4.42\% (detailed performances are not presented to not clutter this section's results).

Next, we trained a VGG-FACE network on all the dimensionally annotated databases to predict valence and arousal; we also trained another VGG-FACE network on all categorically annotated databases, to perform seven basic expression classification; finally we trained a third VGG-FACE network on all databases annotated with action units, so as to perform AU detection. For brevity these three single-task networks are denoted as '(3 $\times$) VGG-FACE single-task' in one row of Table \ref{comparison_sota}.

We compared these networks' performances with the performance of FaceBehaviorNet when trained with and without the coupling losses. We also compare them with the performances of the state-of-the-art methodologies of each utilized database: 
i) FATAUVA-Net \cite{weichi} (described in Section \ref{related_work}), which was the winner of Aff-Wild Challenge; 
ii) the best performing CNN (VGG-FACE) on Aff-Wild \cite{kollias}\cite{kollias2018deep}; 
iii) the baseline networks (AlexNet) on AffectNet  \cite{mollahosseini2017affectnet} (in Table \ref{comparison_sota} they are denoted as '(2 $\times$) AlexNet' as they are two different networks: one for VA estimation and another for expression classification); 
iv)  the baseline network (non-linear Chi-square kernel based SVM)  \cite{dhall2018emotiw} on EmotiW Challenges; 
v) VGG-FACE-mSVM \cite{li2017reliable} on RAF-DB; 
vi)  the baseline network (AlexNet) on EmotioNet  \cite{benitez2017emotionet};
vii) ResNet-34, which was the best performing network on EmotioNet \cite{ding2017facial};
viii) Discriminant Laplacian Embedding extension (DLE extension)\cite{yuce2015discriminant}, which was the winner of FERA 2015 on BP4D;
ix) \cite{tang2017view}, which was the winner of FERA 2017 on BP4D+. Table \ref{comparison_sota} displays the performances of all these networks.

Here, let us mention that in Aff-Wild the best performing network is AffWildNet \cite{kollias2018deep} \cite{kollias}, that has a CCC of 0.57 and 0.43 in valence and arousal respectively; this network is a CNN-RNN that exploits the fact that the Aff-Wild database is an audio-visual one. Additionally,  facial landmarks were provided as additional inputs to this network, thus improving its performance. The latter is not included in Table \ref{comparison_sota}. However, although being a CNN-RNN network, its average CCC is the same as the average CCC of our CNN network, FaceBehaviorNet trained with the two coupling losses (in both task relatedness settings). 

Let us also mention that on RAF-DB  the best performing network is the Deep Locality-preserving CNN (DLP-CNN) of \cite{li2017reliable} with a performance metric value of 0.74; this network was trained using a joint classical softmax loss - which forces different classes to stay apart - and a newly created loss - that pulls the locally neighboring faces of the same class together. For the task of expression recognition, our approach used the standard cross entropy loss; therefore a fair comparison cannot be made with our model because DLP-CNN uses a different cost function that we do not use and thus DLP-CNN is not listed in Table \ref{comparison_sota}.

 It might be argued that the more data used for network training (even if they contain partial or non-overlapping annotations), the better network performance will be in all tasks. However this may not  be true, as the three studied tasks are non-homogeneous and each one of them contains ambiguous cases: i) there is generally discrepancy in the perception of the disgust, fear, sadness and (negative) surprise emotions across different people and across databases; ii) the exact valence and arousal value for a particular affect is also not consistent among databases; iii) the AU annotation process is a hard to do and error prone one.
Nevertheless, from Table \ref{comparison_sota}, it can be verified that FaceBehaviorNet achieved a better performance on all databases than the independently trained VGG-FACE single-task models. This shows that, all described facial behavior understanding tasks  are coherently correlated to each other; training an end-to-end architecture with heterogeneous databases simultaneously, therefore, leads to improved performance.

In Table \ref{comparison_sota}, it can be observed that FaceBehaviorNet trained with  no coupling loss: i) ouperforms the state-of-the-art by 3.5\% (average CCC) on Aff-Wild, 4\% (average CCC) on AffectNet, 9\% on RAF-DB and 2\% on BP4D; ii) has the same performance on AFEW; iii) shows inferior performance by 4\% on AffectNet and 1.5\% (on average) on EmotioNet, 1\% on BP4D+. However, when FaceBehaviorNet is trained with soft co-annotation and distr-matching losses (either when task relatedness is inferred from Aff-Wild2 or from \cite{du2014compound}), it shows superior performance to all state-of-the-art methods. The fact that it outperforms these methods and the single-task networks, in both task relatedness settings, verifies the generality of the proposed losses; network performance is boosted independently of the Table of task relatedness which was used.

\subsection{Results: Zero-Shot and Few-Shot Learning}

\noindent In order to further prove and validate that FaceBehaviorNet learned good features encapsulating all aspects of facial behavior, we conducted zero-shot learning experiments for classifying compound expressions.   
Given that there exist only 2 datasets (EmotioNet and RAF-DB) annotated with compound expressions and that they do not contain a lot of samples (less than 3,000 each), at first, we used the predictions of FaceBehaviorNet together with the rules from \cite{du2014compound} to generate compound emotion predictions. Additionally, to  demonstrate  the  superiority  of FaceBehaviorNet, we used it as a pre-trained network in a few-shot learning experiment. We took advantage of the fact that our network has learned good features and used them as priors for fine-tuning the network to perform compound emotion classification.

\paragraph{RAF-DB database}

At first, we performed zero-shot experiments on the 11 compound categories of RAF-DB. We computed a candidate score,  $\mathcal{C}_{s}(y_{emo})$, for each class $y_{emo}$: 
%which was: 

%\smallskip
%\noindent
%$ [\sum_{k=1}^{17} p(y_{au}^k| y_{emo})] ^ {-1} \cdot
%\sum_{k=1}^{17} p(y_{au}^k|x) \: p(y_{au}^k| y_{emo})$ +  

%\smallskip
%\noindent
%$p(y_{emo1}) + p(y_{emo2})$ + $0.5 \cdot (\frac{p(y_{v}|x)}{|p(y_{v}|x)|} + 1)$, $p(y_{v}|x) \neq 0,$

\begin{align}
 \mathcal{C}_{s}(y_{emo}) &= [\sum_{k=1}^{17} p(y_{au}^k| y_{emo})] ^ {-1} \cdot  \sum_{k=1}^{17} p(y_{au}^k|x) \: p(y_{au}^k| y_{emo}) 
\nonumber\\
  &+ p(y_{emo1}) + p(y_{emo2}) 
\nonumber\\
 &+ 0.5 \cdot (\frac{p(y_{v}|x)}{|p(y_{v}|x)|} + 1),     p(y_{v}|x) \neq 0,  
\nonumber
\end{align}

%\smallskip
\noindent
where: i) the first term of the sum is
 FaceBehaviorNet's predictions of only the prototypical (and observational) AUs that are associated with this compound class according to \cite{du2014compound}; in  this  manner,  every AU  acts  as  an  indicator  for  this  particular  emotion  class; this  terms  describes  the  confidence  (probability)  of  AUs that this compound emotion is present; ii) $p(y_{emo1})$ and $p(y_{emo2})$ are FaceBehaviorNet's predictions of only the basic expression classes $emo1$ and $emo2$ that are mixed and form the compound class (e.g., if the compound class is happily surprised then $emo1$ is happy and $emo2$ is surprised); iii) the last term of the sum is added only to the happily surprised and happily disgusted classes and is either 0 or 1 depending on whether FaceBehaviorNet's valence prediction is negative or positive, respectively; the rationale is that only happily surprised and (maybe) happily disgusted classes have positive valence; all other classes are expected to have negative valence as they correspond to negative emotions. 
 Our final prediction was the class that had the maximum candidate score.

Table \ref{comparison_zero_few_shot} shows the results of this approach when we used the predictions of FaceBehaviorNet trained with and without the soft co-annotation and distr-matching losses. Best results have been obtained when the network was trained with the coupling losses. One can observe, that this approach outperformed by 4.8\% the VGG-FACE-mSVM \cite{li2017reliable} which has the same architecture as our network and it has been trained for compound emotion classification. 

Next, we target few-shot learning. In particular, we fine-tune the FaceBehaviorNet (trained with and without the soft co-annotation and distr-matching losses) on the small training set of RAF-DB. In Table \ref{comparison_zero_few_shot} we compare its performance to a state-of-the-art network. It can be seen that our fine-tuned FaceBehaviorNet, trained with and without the coupling losses, outperformed by 1.2\% and 3.7\%, respectively, the best performing network, DLP-CNN, that was trained with a loss designed for this specific task.

\begin{table*}[ht]
\caption{Performance evaluation of generated compound emotion predictions on EmotioNet and RAF-DB databases.
}
\label{comparison_zero_few_shot}
\centering
\scalebox{1.}{
\begin{tabular}{ |c||c|c|c| }
 \hline
\multicolumn{1}{|c||}{\begin{tabular}{@{}c@{}} Databases \end{tabular}} & \multicolumn{2}{c|}{EmotioNet} & \multicolumn{1}{c|}{RAF-DB}  \\
 \hline
 Methods &\multicolumn{1}{c|}{\begin{tabular}{@{}c@{}}  F1 \\ Score \end{tabular}} &  \multicolumn{1}{c|}{\begin{tabular}{@{}c@{}} Unweighted \\ Average Recall \end{tabular}} &\multicolumn{1}{c|}{\begin{tabular}{@{}c@{}}  Mean diagonal \\ of conf. matrix \end{tabular}}   \\
  \hhline{=:=:=:=}
\begin{tabular}{@{}c@{}}  zero-shot, FaceBehaviorNet, no coupling loss \end{tabular} & 0.243 & 0.260  &  0.342   \\
\hline
\begin{tabular}{@{}c@{}} \textbf{zero-shot,  FaceBehaviorNet, both coupling losses}  \end{tabular} & \textbf{0.312}  & \textbf{0.329}   &  \textbf{0.364}   \\
\hline
\hline
NTechLab \cite{benitez2017emotionet} & 0.255 & 0.243 & - \\
\hline
VGG-FACE-mSVM \cite{li2017reliable} & - & - & 0.316 \\
\hline
DLP-CNN \cite{li2017reliable} & - & - & 0.446 \\
\hline
\hline
\begin{tabular}{@{}c@{}} fine-tuned FaceBehaviorNet, no coupling loss \end{tabular} & - & -  &  0.458   \\
\hline
\begin{tabular}{@{}c@{}} \textbf{fine-tuned FaceBehaviorNet, both coupling losses} \end{tabular} & - & -  & \textbf{0.483}    \\
\hline
\end{tabular}
}
\end{table*}

\paragraph{EmotioNet database}

Next, we performed zero-shot experiments on the EmotioNet basic and compound set that was released for the related Challenge. This set includes 6 basic plus 10 compound categories, as described at the beginning of this Section. Our zero-shot methodology was similar to the one described above for the RAF-DB database. 

The results of this experiment can be found in Table \ref{comparison_zero_few_shot}. 
Best results have also been obtained when the network was trained with the two coupling losses. It can be observed that this approach outperformed by 5.7\% and 8.6\% in F1 score and Unweighted Average Recall (UAR), respectively, the state-of-the-art NTechLab's \cite{benitez2017emotionet} approach, which used the Emotionet's images with compound annotation.

\section{Conclusions}

In this paper, we presented FaceBehaviorNet, the first holistic framework for emotional  behaviour  analysis  in-the-wild. FaceBehaviorNet is an end-to-end network trained for joint: basic expression recognition, action unit detection and valence-arousal estimation. All publicly available databases, containing over 5M images, that study facial behaviour tasks in-the-wild, have been utilized. Additionally  we  proposed  two  simple  strategies  for  coupling the  tasks  during  training, namely co-annotation  and  distribution matching. We performed experiments comparing the performance of FaceBehaviorNet to single-task networks, as well as state-of-the-art methodologies. FaceBehaviorNet consistently outperformed all of them. Finally, we explored the feature representation learned in the joint training and showed its generalization abilities on the task of compound expressions, under zero-shot or few-shot learning settings.

%-------------------------------------------------------------------------

{\small
\bibliographystyle{ieee}    
\bibliography{egbib}
}

\end{document}